\documentclass{article}

\usepackage{arxiv}

\usepackage[utf8]{inputenc} 
\usepackage[T1]{fontenc}    
\usepackage{hyperref}       
\usepackage{url}            
\usepackage{booktabs}       
\usepackage{amsfonts}       
\usepackage{nicefrac}       
\usepackage{microtype}      
\usepackage{lipsum}
\usepackage{graphicx}
\graphicspath{ {./images/} }

\title{A Tale of Fairness Revisited:\\ 
Beyond Adversarial Learning for Deep Neural Network Fairness}

\author{
 Mashaido Becky \\
  Khoury College of Computer Sciences \\
  Northeastern University \\
  Boston, MA 02115 \\
  \texttt{mashaido.r@northeastern.edu} \\
   \And
 Winston Moh Tangongho \\
  Khoury College of Computer Sciences \\
  Northeastern University \\
  Boston, MA 02115 \\
  \texttt{mohtangongho.w@northeastern.edu} \\
 
}

\begin{document}
\maketitle
\begin{abstract}
Motivated by the need for fair algorithmic decision making in the age of automation and artificially-intelligent technology, this technical report provides a theoretical insight into adversarial training for fairness in deep learning. We build upon previous work in adversarial fairness, show the persistent tradeoff between fair predictions and model performance, and explore further mechanisms that help in offsetting this tradeoff. 
\end{abstract}


\section{Introduction and Motivation}
\subsection{What is the problem?}
Machine Learning is increasingly becoming a huge part of our lives by automating decision making processes such as who gets a loan\footnote{ Swarns, R. L. Biased lending evolves, and blacks face trouble getting mortgages. The New York Times, 2015.
} or how likely an offender is bound to repeat a crime. Such algorithms however also lack transparency, interpretability and fairness among different user demographics. We address this fairness problem using adversarial learning techniques and explore further mechanisms to help boost the overall performance of our adversary.

\subsection{Why is it interesting and important?}
Depending on one’s incentives and what drives them as a researcher, scientist or developer:

\textbf{Money:} one would not want to build a product that doesn’t live up to its potential in terms of generating income due to its biased targeting. For example, why build a facial recognition system for human beings that services only one demographic and excludes the rest?

\textbf{Morals:} one would want to be responsible for what they build or implement and not let inherent biases facilitate the existing forms of marginalization towards minority groups. For example, if building an applicant tracking system you would ensure that it does not discriminate against women searching for engineering job opportunities

\textbf{Science:} if one seeks to solve a problem for all human beings but fail at including the different demographics, then they achieve wrong scientific results

\subsection{Why is it hard? (E.g., why do naive approaches fail?)}
\renewcommand\labelitemii{$\blacksquare$}
\begin{itemize}
\item Fairness is not well-defined from a computational perspective. Researchers don’t have a common understanding on how to define, measure and address fairness
\item Fixing models for fairness is also not as straightforward as removing race and sex attributes from training data since models can still pick up on other biases that correlate with such protected attributes. For example, you may remove the race attribute from a dataset but your model remains unfair since it learns the zip code attribute, which tends to go hand in hand with racial geographical demographics 
\item Naive approaches attempt to explicitly train the model with adversarial examples to improve generalization of the model but this approach does not make the model completely robust to unseen adversarial examples
\item Fairness is broad and has to be applied to respective domains, it is not just one approach applied to every domain in the same manner. You have to tweak it appropriately, to the respective domain, in order to achieve overall fairness i.e not easily transferable 
\item Fairness tends to be an afterthought among machine learning researchers, instead of incorporating it throughout the process
\item There is not enough collaboration between technical and social researchers
\end{itemize}

\subsection{Why hasn’t it been solved before? (Or, what’s wrong with previous proposed solutions? How does mine differ?)}
In addition to points laid out in section 1.3, previous proposed solutions tend to sacrifice a bit of model accuracy in order to achieve fairness. We however try to improve both fairness and accuracy by exploring further mechanisms such as the gradient descent-ascent algorithm, with a modified gradient update step, inspired by previous work \cite{art1}. This is all in addition to adversarial training

We initially design our model following the work done by \cite{art3} and replicate the work done by \href{https://godatadriven.com/blog/towards-fairness-in-ml-with-adversarial-networks/}{Go Data Driven} on a COMPAS Dataset 

Finally, we show that with some improvements to the gradient update step we can substantially improve our model’s fairness, while maintaining good model accuracy

Past research has explored the use of adversarial networks to achieve fairness on the COMPAS dataset. Some have used the metrics of false positive and false negative differences across Black and White inmates. One such example \cite{art2} used post-processing on a standard logistic regressor that picks different thresholds for different groups and at times added randomization. Another \cite{art5} introduced penalties for misclassified data points during training to enforce equal false positive and false negative rates for groups. This approach then performed some optimization using the covariance between sensitive attributes and distance between feature vectors of misclassified samples and classifier boundary. Another approach \cite{art4} demonstrated how one can use adversarial learning on real-world applications. We therefore expand on their work by using a modified gradient update step.

\subsection{What are the key components of my approach and results? Also include any specific limitations.}

We use an \cite{art1} adversarial approach on the COMPAS dataset. We aim to predict the chances of recidivism of inmates at the Broward County jail in Florida, and apply some adversarial learning approaches to optimize our model for fairness. We look at the COMPAS model’s score and probe it for racial bias. For this project, we set our sensitive attribute as race (Caucasian or African-American). 

We define our fairness metric as a statistical rate, with the following inequality:

\begin{equation}
min(\frac{P(y\hat{} = 1 | z = 1)}{P(y\hat{} = 1 | z = 0)}, \frac{P(y\hat{} = 1 | z = 0)}{P(y\hat{} = 1 | z = 1)}) \geq \frac{p}{100}
\end{equation} 

This rule states that the ratio between the probability of a positive outcome given the sensitive attribute being true and the same probability given the sensitive attribute being false is no less than p:100. A completely fair classifier will satisfy a 100\% rule and a completely unfair one will satisfy a 0\% rule.
For this project, we use the U.S. Equal Employment Opportunity (EEOC) and their 80\% rule to quantify the disparate impact on a group of people of a protected characteristic.

We then perform experiments and observe results. If we obtain both fairness and good model accuracy, well and good. If we obtain fairness but not good model accuracy then we explore further methods such as using a modified gradient update step. If we obtain good model accuracy but not achieving fairness, then we counteract scientists’ claims that adversarial helps improve fairness in deep learning.

\textbf{Specific limitations:} In the case that we achieve both fairness and good model accuracy it is not a definite indicator that this knowledge transfers well to other models or tasks in adversarial environments.

\section{Adversarial Model and Methods}
Assuming a probability model $P(X,Y,Z)$, where $X$ are the data, $Y$ are the target labels, and $Z$ are the nuisance parameters (sensitive attributes), we consider the problem of learning a predictive model $f(X)$ for $Y$ conditional on the observed values of X that is robust to uncertainty in the unknown value of $Z$ \cite{art3}.

\begin{figure}[htp]
    \centering
    \includegraphics[scale=0.5]{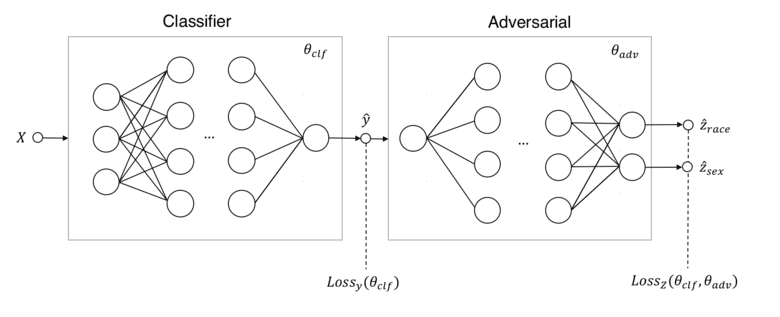}
    \caption{Architecture for the adversarial training of a binary classifier against a sensitive parameters $Z$. The adversary maximizes its objective $Loss_{z}(\theta_{clf}, \theta_{adv})$ while the classifier tries to minimize its objective $Loss_{y}(\theta_{clf})$}
\end{figure}

We replicate the work from \href{https://godatadriven.com/blog/towards-fairness-in-ml-with-adversarial-networks/}{Go Data Driven} where for the classifier we try to make the best possible recidivism prediction while ensuring that race cannot be derived from them. This is captured by the following objective function:

\begin{equation}
min_{\theta_{clf}}[Loss_{y}(\theta_{clf}) - \lambda Loss_{Z}(\theta_{clf}, \theta_{adv})]
\end{equation} 

It learns to minimize its own prediction losses while maximizing that of the adversarial (due to $\lambda$ being positive and minimizing a negated loss is the same as maximizing it). The adversary learns on the full dataset and the classifier is given only the single batch, giving the adversary a slight edge in learning.

For the adversary, its goal is to predict the race based on the recidivism score of the classifier. It is captured in the following objective function:

\begin{equation}
min_{\theta_{adv}}[Loss_{Z}(\theta_{clf}, \theta_{adv})]
\end{equation}

\subsection{Adversarial Training Procedure}
\begin{enumerate}
\item    Pre-train the classifier on the full data set
\item Pre-train the adversarial on the predictions of the pre-trained classifier
\item During $T$ iterations simultaneously train the adversarial and classifier networks:
    \begin{enumerate}
    \item First train the adversarial for a single epoch while keeping the classifier fixed
    \item Then train the classifier on a single sampled mini batch while keeping the adversarial fixed
    \end{enumerate}
\end{enumerate}

Note that the actual adversarial training starts only after the first two pre-training steps.
The results after applying this procedure are shown in Figure 3.

\subsection{ Improvement with Regularized Logistic Regression}
The work done by \cite{art1} shows that with their improved gradient descent algorithms, we can create models that perform comparably or have better accuracy than other algorithms while performing better in terms of fairness, as measured by using statistical rate (p\% value). In this project, we explore two of their algorithms described in the paper and test them on our COMPAS dataset.

For a given weight vector $w \in R^{n+1}, f = \sigma (w^{T}x\hat{})$, where $\sigma(\cdot)$ is the sigmoid function. We then add some noise to the inputs to the classifier to make it partially randomized. This improves our model accuracy if the gradient between the gradient of the classifier and gradient of the adversary are highly correlated. We accomplish this by multiplying the inputs by some noise scalar $\eta$ uniformly chosen from $[0,1]$.

Next, we update the Loss functions for the classifier and the adversary to obtain:

\begin{figure}[htp]
    \centering
    \includegraphics[scale=0.5]{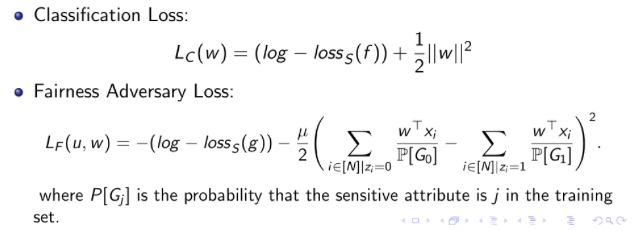}
\end{figure}

The first part of the Adversary loss function corresponds to learning the correlation between the sensitive attribute and the class label. The second part of the loss function is a regularizer to check whether fairness is satisfied.
Having defined our loss functions, we are ready to introduce the gradient descent algorithms.

\textbf{i) Gradient descent/ascent with Normal Update:}  Here, we use Normal gradient descent/ascent algorithm to minimize $L_{C}$ and maximize $L_{F}$.By combining the two directions with a controlling parameter $\alpha > 0$, we get:

\begin{equation}
u_{t+1} = u_{t} + \eta_{1} \nabla_{u} L_{F}, 
\end{equation}
\begin{equation}
w_{t+1} = w_{t} - \eta_{2} (\nabla_{w} L_{C} - \alpha \cdot \nabla_{w} L_{F})
\end{equation}

for some $\eta_{1}$, $\eta_{2}$, $\alpha$ $>$ $0$.\\

We see the results of this algorithm in Figure 4.\\

\textbf{ii) Normal Gradient Descent with Modified Update:}  In the event that the gradient of fairness loss and gradient of classification loss are highly correlated, the previous algorithm might not ensure both fairness and accuracy. In this algorithm, we remove the projection of $\nabla_{w} L_{F}$ from $\nabla_{w} L_{C}$ from the update. At each iteration we use the following steps:

\begin{equation}
u_{t+1} = u_{t} + \eta_{1} \nabla_{u} L_{F}, 
\end{equation}
\begin{equation}
w_{t+1} = w_{t} - \eta_{2} (\nabla_{w} L_{C} - \alpha \cdot \nabla_{w} L_{F} - \Pi_{\nabla_{w} L_{F}} \nabla_{w} L_{C})
\end{equation}

for some $\eta_{1}$, $\eta_{2}$, $\alpha$ $=$ $\frac{1}{\sqrt{t}}$.\\

Since the problem is a multi-objective optimization problem, we do not expect it to converge to the optimal point after T iterations and so we apply thresholding. We therefore record the parameters for which the statistical rate and the training accuracy is maximum and use it as a way to control the fairness of the output classifier. Our results are shown in Figure 5.

\section{Experiments and Results}
We evaluate the performance of our methods empirically and report the classification accuracy and fairness on the COMPAS dataset. We also compare the different update methods.

\subsection{Dataset}
COMPAS (Correctional Offender Management Profiling for Alternative Sanctions) is a popular commercial algorithm used by judges and parole officers for scoring criminal defendant’s likelihood of reoffending (recidivism). It has been shown that the algorithm is biased in favor of white (against black) inmates based on a 2 year follow up study on who actually committed crimes or violent crimes after 2 years. 

We used the non violent subset of the data for offenses such as vagrancy or marijuana. So we dropped the violent offense attributes from our table.

\subsection{Experiments}
For all experiments, code was written in pytorch, Dataset was the COMPAS dataset with 15201 samples and 42671 attributes (one-hot encoded). A Neural network consisting of 3 hidden layers with 32 hidden neurons at each layer was used. Maximum number of epochs was 165. 

\textbf{Note:} Legend of plots: African-American class is represented by the blue distribution and the Caucasian class is represented with the pink distribution.

Before we start adversarial training, we pre-train the classifier and our adversary and notice the need for fairness optimization. We can observe the results in Figure 2.

\begin{figure}[htp]
    \centering
    \includegraphics[scale=0.5]{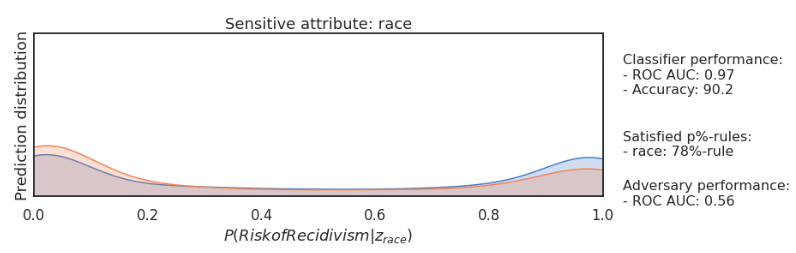}
    \caption{Pre-trained classifier and adversary. The plot shows a higher risk of recidivism for black inmates characterized by the blue colored distribution. Also, the p\% is below 80 showing a hint of bias and with the Adversary ROC AUC above 0.5, it has the ability to detect race from the output of the classifier.}
\end{figure}

\begin{figure}[htp]
    \centering
    \includegraphics[scale=0.5]{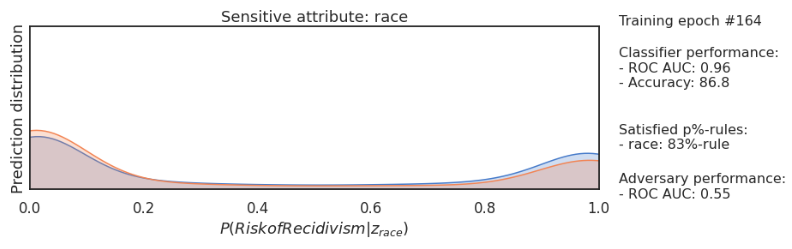}
    \caption{After applying the Adversarial Training procedure. We notice the improvement in ROC AUC to 0.55 which is still above 0.50 and the p\% increase to 83. We have a sharp drop in classifier accuracy and so we explore other algorithms to mitigate this.}
\end{figure}

\begin{figure}[htp]
    \centering
    \includegraphics[scale=0.5]{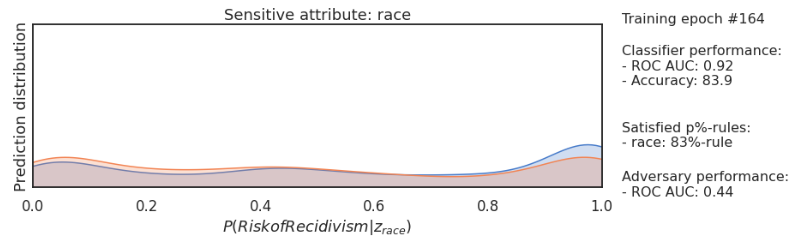}
    \caption{GD (Gradient Descent) with Normal Update. Massive improvement in Model fairness as seen by the low Adversary AUC of 0.44. Disadvantage is massive drop in accuracy to 83.9.}
\end{figure}

\begin{figure}[htp]
    \centering
    \includegraphics[scale=0.5]{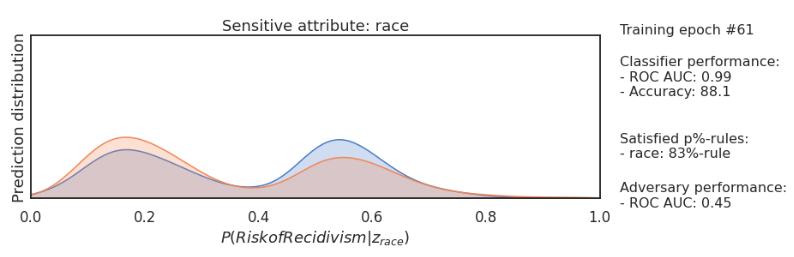}
    \caption{ Normal GD with Modified Update. Here the threshold =61. The Model accuracy is increased to 88.1 while maintaining fairness.}
\end{figure}

\section{Conclusion}
We have shown how adversarial learning can be used to improve fairness in deep neural networks and made observations on the persistent tradeoff between fairness and performance. We further demonstrated that we can offset this tradeoff by using the normal gradient descent algorithm with a modified update step.

Our project uses adversarial learning to ensure fairness, though there is scope for further research. We started by applying a gradient descent-ascent to train both our adversary and classifier, then improved our algorithm by applying a regularizer to improve fairness of the model. Lastly, we optimized the gradient update step to remove the projection of the gradient of the adversary from the gradient of the classifier. This led us to find an optimally-fair and accurate model. 

Ensuring fairness and accuracy is a difficult task but with the huge traction fairness has gained in both academia and industry, this work is promising. For future work, we propose that adversarial learning be explored in the context of individual fairness, rather than group-fairness metrics, which we considered here.\\

--------------------

\bibliographystyle{unsrt}  


\end{document}